\documentclass[journal, 10pt]{IEEEtran}
\usepackage{epsfig,endnotes}
\usepackage{url}
\usepackage{xcolor,colortbl}
\usepackage{booktabs,caption,fixltx2e}
\usepackage[flushleft]{threeparttable}
\usepackage{wrapfig}
\usepackage{float,graphicx}
\usepackage[font=small]{caption}
\usepackage{subcaption}
\usepackage{multicol}
\usepackage{multirow}
\usepackage{float}
\usepackage{tikz}
\usepackage{mwe}
\usepackage{listings}
\usepackage{amsmath,mathtools,amssymb}
\usepackage{array, makecell}
\usepackage{enumitem}
\usepackage{microtype}
\usepackage{verbatim}
\usepackage{setspace}
\usepackage{bookmark}
\usepackage[ruled,linesnumbered]{algorithm2e}

\SetCommentSty{mycommfont}
\usepackage{xcolor,colortbl}
\definecolor{ashgrey}{rgb}{0.7, 0.75, 0.71}
\SetKw{KwBy}{by}
\usepackage{textcomp}
\makeatletter
\g@addto@macro{\UrlBreaks}{\UrlOrds}

\definecolor{Gray}{gray}{0.85}
\definecolor{lblue}{rgb}{0.7,1,1}
\usepackage[belowskip=-15pt,aboveskip=0pt]{caption}

\definecolor{codegray}{rgb}{0.5,0.5,0.5}
\definecolor{codepurple}{rgb}{0.58,0,0.82}
\definecolor{backcolour}{rgb}{0.95,0.95,0.92}

 \lstdefinestyle{mystyle}{
    backgroundcolor=\color{backcolour},   
    commentstyle=\color{codegray},
    keywordstyle=\color{magenta},
    numberstyle=\tiny\color{codegray},
    stringstyle=\color{codepurple},
    basicstyle=\ttfamily\scriptsize,
    breakatwhitespace=false,         
    breaklines=true,                 
    captionpos=b,                    
    keepspaces=true,                               
    numbersep=5pt,                  
    showspaces=false,                
    showstringspaces=false,
    showtabs=false,                  
    tabsize=2
}

\makeatletter
\def\endthebibliography{%
  \def\@noitemerr{\@latex@warning{Empty `thebibliography' environment}}%
  \endlist
}
\ifCLASSINFOpdf
\else
\fi

\makeatother

\begin{document}

\title{Linear Feedback Control Systems for Iterative Prompt Optimization in  Large Language Models\vspace{-5pt}}

\author{
  \IEEEauthorblockN{Rupesh Raj Karn}\\
  \IEEEauthorblockA{Center for Cyber Security, New York University, Abu Dhabi, UAE.}\\
  Email: rupesh.k@nyu.edu
}

\maketitle

\begin{abstract}
Large Language Models (LLMs) have revolutionized various applications by generating outputs based on given prompts. However, achieving the desired output requires iterative prompt refinement. This paper presents a novel approach that draws parallels between the iterative prompt optimization process in LLMs and feedback control systems. We iteratively refine the prompt by treating the deviation between the LLM output and the desired result as an error term until the output criteria are met. This process is akin to a feedback control system, where the LLM, despite being non-linear and non-deterministic, is managed using principles from linear feedback control systems. We explore the application of different types of controllers within this framework, providing a mathematical foundation for integrating linear feedback control mechanisms with LLMs. 
\end{abstract}
\IEEEpeerreviewmaketitle



\vspace{-10pt}
\section{Introduction}
\label{sec:introduction}

Large Language Models (LLMs) have become influential tools across various fields, such as natural language processing, automated code generation, and others \cite{hadi2024large}. These models produce outputs based on provided prompts, but achieving the desired outcomes necessitates iterative prompt refinement \cite{dai2024assessing}. This iterative process resembles a feedback control system, where the difference between the generated output and the desired result is considered an error term. By continuously refining the prompt, we strive to minimize this error and achieve the desired result \cite{ouyang2022training}.

The concept of feedback control systems is well-established in control theory, where it is used to manage the behavior of dynamic systems \cite{ozbay2019introduction}. In a typical feedback control system, the system's output is continuously monitored and compared to the desired output \cite{aastrom2021feedback}. Any deviation (error) is fed back into the system to adjust the inputs, thereby reducing the error over time. This paper proposes a novel approach to apply these principles to optimize LLM outputs.

Despite the non-linear and non-deterministic nature of LLMs \cite{ouyang2023llm}, we explain that principles from linear feedback control systems \cite{zhou2014optimal} can be effectively applied to optimize their outputs. By integrating different types of controllers, such as Proportional-Integral-Derivative (PID) controllers \cite{johnson2005pid}, we provide a structured methodology for enhancing the performance and reliability of LLM-driven applications. Traditional methods of prompt optimization often rely on heuristic or trial-and-error approaches, which can be time-consuming and inefficient \cite{beurer2023prompting}. By leveraging the systematic approach of feedback control systems, we aim to provide a more robust and theoretically grounded method for prompt optimization, thereby advancing the state-of-the-art in automated prompt refinement. \textit{The follow-up publication will detail the practical implementation and evaluation of this theory across diverse computing applications and release the source code for AI community.}

This paper makes the following key contributions:
\begin{enumerate}
    \item Introduces a novel framework that applies linear feedback control system principles to the iterative prompt optimization process in LLMs.
    \item Provides a detailed mathematical foundation for integrating PID feedback controller mechanisms with LLMs and displays its functionality with an FPGA design example.
\end{enumerate}

\vspace{-5pt}
\section{Preliminaries: Feedback Control System}
\label{subsec:preliminaries}
\begin{figure}[!t]
	\begin{center}
		\includegraphics[scale=0.7, trim = {0.7cm 0 0 0}, clip]{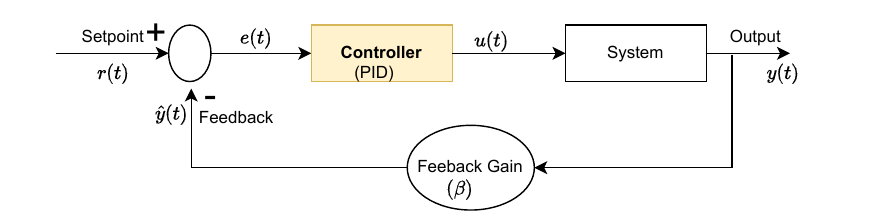} 
		\caption{General feedback control system.} 
		\label{fig:feedbackloop_diag}
	\end{center}
 \vspace{-10pt}
\end{figure}

A feedback control system shown in Fig.~\ref{fig:feedbackloop_diag} is a dynamic system that automatically adjusts its operation to meet a reference point \cite{borase2021review}. The controller processes the error signal and generates a control action to minimize this error, thereby driving the system towards the desired performance. 

A Proportional-Integral-Derivative (PID) controller is a widely used feedback control mechanism in industrial automation and control systems \cite{kumar2021diffloop}. It aims to regulate a process variable by adjusting a manipulated variable based on the error between the setpoint and the actual process variable.

The PID controller combines three control actions: proportional, integral, and derivative. The control output \( u(t) \) is given by:
\begin{equation}
\label{equ:pid_controller}
u(t) = K_p e(t) + K_i \int_{0}^{t} e(\tau) d\tau + K_d \frac{de(t)}{dt}
\end{equation}
where:
\begin{itemize}[label=$\triangleright$]
    \item \( e(t) \) is the error signal, \( y(t) \): \( e(t) = r(t) - \hat{y}(t) \).
    \item \( K_p, K_i, K_d \) are the proportional, integral, and derivative gains, respectively.
\end{itemize}
The proportional term provides an output that is proportional to the current error value $e(t)$. The integral term accounts for the accumulation of past errors. It integrates the error over time to eliminate the steady-state error. The derivative term predicts the error's future trend by considering its change rate. It provides a damping effect, reducing overshoot and oscillations. Combining the three control actions, the overall PID control law is expressed as in equ~(\ref{equ:pid_controller}).

The error signal \( e(t) \) is fed into the PID controller, which computes the control output \( u(t) \) to adjust the process variable \( \hat{y}(t) \) to match the setpoint \( r(t) \). The term $\beta$ is feedback gain and  \( \hat{y}(t) = \beta y(t) \).

The performance of a PID controller depends on the proper tuning of its parameters \( K_p \), \( K_i \), and \( K_d \). Various tuning methods, such as the Ziegler-Nichols method \cite{patel2020ziegler}, Cohen-Coon method \cite{borase2021review}, are used to determine these parameters to achieve the desired system performance.

\begin{figure}[!t]
	\begin{center}
		\includegraphics[scale=0.6, trim = {0.7cm 0 0 0}, clip]{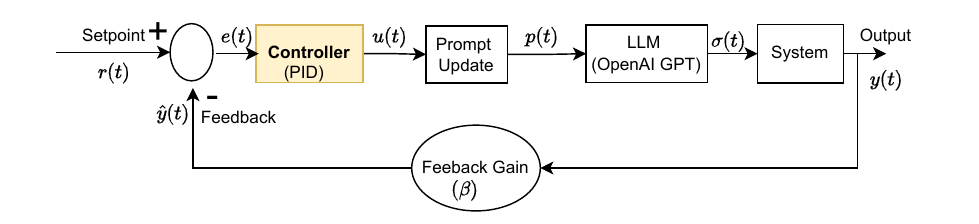} 
		\caption{LLM feedback control system for prompt refinement.} 
		\label{fig:LLM_control_system}
	\end{center}
  \vspace{-10pt}
\end{figure}

\vspace{-10pt}
\section{Feedback Control for LLMs}

\subsection{Overall Architecture}
\label{subsec:overall_archt}
The feedback loop mechanism employed in our approach for iterative prompt optimization using LLMs is illustrated in Fig.~\ref{fig:LLM_control_system}. We integrate principles from linear feedback control systems with modern machine learning techniques employed in LLMs in this setup.

Like Fig.~\ref{fig:feedbackloop_diag}, the control signal \( u(t) \) is calculated through equ~(\ref{equ:pid_controller}). In this case, the \( u(t) \) is used to update the prompt $p(t)$ through the 'Prompt Update (OpenAI GPT \cite{roumeliotis2023chatgpt})' block. 
\begin{equation}
\label{equ:prompt_update}
    p(t+1) = p(t) + u(t)
\end{equation}    
The updated prompt \( p(t+1) \) is subsequently processed by the LLM block (e.g., OpenAI's GPT) to generate the output \( \sigma(t) \), which is then integrated into the system use case. The LLM's output is modeled as:
\begin{equation}
    \sigma(t+1) = f\left(p(t+1)\right)
\end{equation}
Here, the function \( f \) represents the LLM. Similarly, the system function is denoted by \( \phi \), whose output is given as:
\begin{equation}
    y(t+1) = \phi\left(\sigma(t+1)\right)
\end{equation}
The output \( y(t) \) is fed back into the system through a feedback gain \( \beta \), completing the loop.

This feedback loop allows for continuous refinement of the prompt $p(t)$, ensuring that the output \( y(t) \) converges towards the desired setpoint \( r(t) \). Integrating traditional control theory with advanced machine learning techniques applied in LLMs provides a structured methodology for optimizing LLM outputs.

\vspace{-10pt}
\subsection{Incorporating LLM Properties into Feedback Control}
\label{subsec:llm_properties}
LLMs exhibit several key properties, including stochasticity \cite{saba2023stochastic}, non-determinism \cite{naveed2023comprehensive}, and inherent non-linearity \cite{raiaan2024review,lin2024towards}. These properties significantly influence the behavior of the feedback control loop used for iterative prompt optimization. 

\subsubsection{Stochastic and Non-Deterministic Outputs}

LLMs generate outputs that are inherently stochastic and non-deterministic. For a given prompt \( p(t) \), the output \( \sigma(t) \) can vary across different iterations. To model this behavior, we introduce a stochastic noise term \( \eta(t) \) into the LLM output equation:
\begin{equation}
    \sigma(t+1) = f\left(p(t+1)\right) + \eta(t)
\end{equation}
Here, \( \eta(t) \) represents the stochastic variations in the LLM output, which can be modeled as a random variable with a specific probability distribution.

\subsubsection{Non-Linearity in LLM Architecture}

The architecture of LLMs is highly non-linear, which affects the relationship between the input prompt \( p(t) \) and the output \( \sigma(t) \). To capture this non-linearity, we modify the function \( f \) to include a non-linear transformation \( g \):
\begin{equation}
\label{equ:non-linear_sigma}
    \sigma(t+1) = g\left(f\left(p(t+1)\right)\right) + \eta(t)
\end{equation}
The function \( g \) could be activation functions or attention mechanisms that are applied in LLM architecture.

\vspace{-10pt}
\subsection{Refined System Output}

The system output \( y(t) \) is influenced by the non-linear and stochastic nature of the LLM. We modify the output equation to include these effects:
\begin{equation}
\label{equ:non-linear_output}
    y(t+1) = \phi\left(\sigma(t+1)\right) + \nu(t)
\end{equation}
Here, \( \nu(t) \) represents additional noise introduced during the output processing stage, capturing any uncertainties or variations in the final output.

By incorporating the stochastic, non-deterministic, and non-linear properties of LLMs into the feedback loop equations, we enhance the robustness and accuracy of the iterative prompt optimization process. 

\vspace{-5pt}
\section{LLM Output Generation with PID Control}

In this section, we describe the mechanism by which an LLM processes the input prompt \( p(t) \) to generate the output \( \sigma(t) \), incorporating the effects of a PID controller. 

\vspace{-10pt}
\subsection{Tokenization and Embedding}

The first step in processing the updated prompt \( p(t+1) \) involves tokenization \cite{kenton2019bert}, where the input text is divided into smaller units called tokens. Each token is converted into a high-dimensional vector through an embedding layer \cite{vaswani2017attention}. Let \( p(t+1) \) be tokenized into \( \{p_1, p_2, \ldots, p_n\} \). The embedding process can be represented as:
\begin{equation}
\label{equ:tokenization}
    \mathbf{e}_i = \text{Embed}(p_i + u(t)), \quad i = 1, 2, \ldots, n
\end{equation}
where \( \mathbf{e}_i \) is the embedding vector corresponding to the token \( p_i \), and the PID controller output \( u(t) \) influences the tokenization process by adjusting the prompt due to equ~(\ref{equ:prompt_update}).

\vspace{-10pt}
\subsection{Positional Encoding}

To incorporate the order of tokens, positional encoding is added to the embedding vectors \cite{vaswani2017attention}. This can be mathematically expressed as:
\begin{equation}
\label{equ:positonal_encoding}
    \mathbf{e}_i' = \mathbf{e}_i + \mathbf{PE}(i + u(t))
\end{equation}
where, \( \mathbf{PE}(i) \) is the positional encoding vector for the \( i \)-th position, and \( u(t) \) affects the positional encoding by modifying the position indices.

\vspace{-10pt}
\subsection{Transformer Layers}

The core of the LLM consists of multiple transformer layers \cite{patil2024review}, each comprising self-attention \cite{song2024low} and feed-forward sub-layers \cite{patil2024review}. The self-attention mechanism computes a weighted sum of the input embeddings, allowing the model to focus on different parts of the input sequence. The self-attention operation is given by:
\begin{eqnarray}
\nonumber
    \text{Attention}(\mathbf{Q}(u(t)), \mathbf{K}(u(t)), \mathbf{V}(u(t))) =  \\ 
    \label{equ:transformer}
    \text{softmax}\left(\frac{\mathbf{Q}(u(t)) \mathbf{K}(u(t))^T}{\sqrt{d_k}}\right) \mathbf{V}(u(t))
\end{eqnarray}
where \( \mathbf{Q}(u(t)) \), \( \mathbf{K}(u(t)) \), and \( \mathbf{V}(u(t)) \) are the query, key, and value matrices derived from the input embeddings influenced by the PID controller output \( u(t) \).

Each transformer layer also includes a feed-forward network (FFN) applied to each position separately and identically:
\begin{equation}
\label{equ:ffn}
    \text{FFN}(\mathbf{x}(u(t))) = \text{ReLU}(\mathbf{x}(u(t)) \mathbf{W}_1 + \mathbf{b}_1) \mathbf{W}_2 + \mathbf{b}_2
\end{equation}
where \( \mathbf{W}_1 \), \( \mathbf{W}_2 \), \( \mathbf{b}_1 \), and \( \mathbf{b}_2 \) are learnable parameters, and \( u(t) \) affects the input \( \mathbf{x} \).

\vspace{-10pt}
\subsection{Output Generation}

After passing through several transformer layers, the final hidden states are used to generate the output tokens. This involves a linear transformation \cite{din2023jump} followed by a softmax function to produce a probability distribution over the vocabulary:
\begin{equation}
\label{equ:output_generation}
    \mathbf{o}_i = \text{softmax}(\mathbf{h}_i(u(t)) \mathbf{W}_o + \mathbf{b}_o)
\end{equation}
where \( \mathbf{h}_i(u(t)) \) is the hidden state of the \( i \)-th token influenced by \( u(t) \), and \( \mathbf{W}_o \) and \( \mathbf{b}_o \) are the output layer parameters.

The output \( \sigma(t+1) \) is then generated by sampling from the probability distribution \( \mathbf{o}_i \).

\vspace{-10pt}
\subsection{Incorporating PID Control into LLM Output Generation}

The PID controller adjusts the prompt \( p(t) \) iteratively to minimize the error \( e(t) \). This updated prompt \( p(t+1) \) is then processed by the LLM to generate the output \( \sigma(t+1) \). The function \( f \) that represents the LLM's processing can be broken down as follows:
\begin{eqnarray}
\nonumber    f(p(t+1)) = \text{softmax}\left( \text{FFN}\left( \text{Attention}\left( \mathbf{Q}(p(t) + u(t)), \right. \right. \right. \\
        \left. \left. \left. \mathbf{K}(p(t) + u(t)), \mathbf{V}(p(t) + u(t)) \right) \right) \right)    
\end{eqnarray}
Here, \( \mathbf{Q}(p(t) + u(t)) \), \( \mathbf{K}(p(t) + u(t)) \), and \( \mathbf{V}(p(t) + u(t)) \) are the query, key, and value matrices derived from the tokenized and embedded prompt \( p(t+1) \). 

The control signal \( u(t) \) directly affects the embeddings and positional encodings, thereby influencing the self-attention mechanism and the subsequent feed-forward network. This results in the generation of the output \( \sigma(t+1) \) that is optimized based on the PID controller's adjustments.

This detailed breakdown illustrates the complex yet structured process by which LLMs transform input prompts into coherent outputs, leveraging advanced deep learning and mathematical operations while incorporating the adjustments made by the PID controller to optimize the output.

\vspace{-10pt}
\subsection{Impact Analysis}

We analyze the impact of the control signal \( u(t) \) on the various stages of the LLM processing pipeline. The control signal \( u(t) \) is composed of three components: proportional error, integral error, and derivative error, with corresponding gains \( K_p \), \( K_i \), and \( K_d \). We will examine how each component affects the Tokenization and Embedding, Positional Encoding, Transformer Layers, and Output Generation stages.

\subsubsection{Tokenization and Embedding}

The control signal \( u(t) \) affects the tokenization and embedding process by adjusting the prompt \( p(t) \). The proportional component \( K_p e(t) \) provides immediate adjustments to the token embeddings (equ~(\ref{equ:tokenization})), ensuring that the prompt reflects the current error state. The integral component \( K_i \int_{0}^{t} e(\tau) d\tau \) influences the embeddings by incorporating the history of errors, thereby refining the prompt based on accumulated deviations. The derivative component \( K_d \frac{de(t)}{dt} \) helps in anticipating future trends and smoothing the embeddings to prevent abrupt changes.

\subsubsection{Positional Encoding}

In positional encoding, the proportional component directly adjusts the position indices, ensuring immediate alignment with the current error. The integral component corrects long-term deviations in positional encoding, enhancing the model's ability to maintain context over time. The derivative component smooths positional adjustments, reducing oscillations and ensuring stable positional encoding.

\subsubsection{Transformer Layers}

The transformer layers, comprising self-attention and feed-forward networks, are significantly impacted by the control signal \( u(t) \). The proportional component adjusts the query, key, and value matrices in the self-attention mechanism, ensuring the attention weights reflect the current error. The integral component influences the attention mechanism by incorporating past errors, refining the model's focus over time. The derivative component helps predict future changes, smoothing the attention weights to prevent abrupt shifts.

In the feed-forward network (FFN), the proportional component immediately affects the input \( \mathbf{x} \), ensuring that the network responds to the current error. The integral component ensures long-term stability by incorporating the history of errors, while the derivative component smooths the input to the FFN, reducing oscillations.

\subsubsection{Impact Analysis Summary}

Overall, the proportional component \( K_p e(t) \) has the most immediate and significant impact on all stages of the LLM processing pipeline. In contrast, the integral and derivative components ensure long-term stability and smoothness. The combined effect of the control signal \( u(t) \) and the previous prompt \( p(t) \) results in the optimized generation of the output \( \sigma(t+1) \) (equ~(\ref{equ:non-linear_sigma})).

\vspace{-10pt}
\subsection{Session-Based Impact of PID Control}

In cases where the LLM remembers the current session, typically through a GUI interface (e.g., Chat GPT \cite{sharma2022chat}, Google's Gemini \cite{singh2023chat}, Microsoft Co-pilot \cite{stratton2024introduction}), the PID controller significantly enhances performance by leveraging the history of prompts and responses. The integral component accumulates past errors, and the derivative component anticipates future trends, both contributing to a more refined and stable output. For example, in a customer support chatbot integrated into a web application, the LLM can maintain context across multiple interactions within the same session, allowing for more coherent and contextually relevant responses \cite{chin2024human}.

In contrast, LLMs accessed via APIs, such as OpenAI’s GPT models, do not retain history between sessions, as each call creates a new session \cite{liu2023gpt}. For instance, an LLM used in a serverless architecture where each API call is stateless will not remember previous interactions, leading to each request being processed independently \cite{ma2024my}. In such scenarios, the integral and derivative components are effectively zero, as there is no historical data to accumulate or trends to predict. Consequently, the LLM output is directly influenced by the proportional term \( K_p e(t) \), which provides immediate adjustments based solely on the current error.

\vspace{-5pt}
\section{Compare: PID vs Other Controllers for LLM}
We compare the effectiveness of different controllers—PID, Lead-Lag \cite{jadoon2017comparative}, Linear-Quadratic Regulator (LQR) \cite{abdullah2022linear}, and Fuzzy Logic Controller \cite{sharma2020mathematical}—in the context of LLM prompt optimization.
\vspace{-10pt}
\subsection{Lead-Lag Controller}

The Lead-Lag controller \cite{jadoon2017comparative} improves system stability and transient response by adding lead and lag compensations:
\[
u(t) = K \left( \frac{T_1 s + 1}{T_2 s + 1} \right) e(t)
\]
Where \( T_1 \) and \( T_2 \) are the lead and lag time constants, respectively. The Lead-Lag controller can enhance the phase margin and improve the transient response but may not handle steady-state errors as effectively as the PID controller due to the lack of integral action.

\vspace{-10pt}
\subsection{Linear-Quadratic Regulator (LQR)}

The LQR \cite{abdullah2022linear} controller minimizes a cost function to achieve optimal control:
\[
J = \int_{0}^{\infty} (x^T Q x + u^T R u) \, dt
\]
\[
u(t) = -K x(t)
\]
Where \( Q \) and \( R \) are weighting matrices, and \( K \) is the gain matrix. The LQR controller provides optimal control by balancing state and control efforts. However, unlike PID, it requires a precise model of the system dynamics, which is hard to compute for LLMs.

\vspace{-10pt}
\subsection{Fuzzy Logic Controller}

The Fuzzy Logic Controller \cite{sharma2020mathematical} uses fuzzy logic to handle uncertainties and non-linearities:
\[
u(t) = \text{Fuzzy}(e(t), \frac{de(t)}{dt})
\]
The Fuzzy Logic Controller can effectively handle non-linearities and uncertainties in LLMs but requires the design of fuzzy rules and membership functions. These can be more complex and time-consuming than PID, making them less efficient for iterative prompt optimization.

The PID controller's proportional, integral, and derivative terms ensure immediate, long-term, and predictive adjustments to the prompt, resulting in a stable and coherent LLM output.

\vspace{-10pt}
\section{Use Case: FPGA Design}

To illustrate the application of our feedback control approach to LLMs, we consider a use case in the domain of neural network implementation on FPGA \cite{bjerge2021scalable,shahshahani2020framework}. The objective is to iteratively refine the prompt ($p(t)$) to generate an optimal design ($y(t)$) that closely matches the desired setpoint ($r(t)$). Specifically, we aim to ensure that the utilization of FPGA resources $\lambda_i, i \in$ \{LUTs, FFs, DSPs, BRAMs\} \cite{kavun2019efficient} for the neural network design is less than setpoint $r(t)$ while meeting timing constraints (positive setup/hold slack time \cite{karn2023securing,karn2024code}).

\vspace{-10pt}
\subsection{Iterative Prompt Refinement Process}

The iterative prompt refinement process involves the following steps:

\begin{enumerate}
    \item \textit{Initial Prompt:} Start with a prompt \( p(0) \) to generate the HLS C code for the neural network implementation.
    \item \textit{LLM Output:} The LLM processes the prompt and generates the initial design specification \( \sigma(0) \) as the HLS C code, represented by equ~(\ref{equ:non-linear_sigma}).
    \item \textit{Resource Utilization Calculation:} The generated design is synthesized using tools like Vivado \cite{o2014xilinx}, Quartus \cite{snider2022chapter}, to obtain the utilization values of LUTs, FFs, DSPs, BRAMs, and timing slack. This is represented through equ~(\ref{equ:non-linear_output}). The utilization \( y(0) \) is calculated for each resource. Also, assume the feedback gain $\beta=1$.
    \item \textit{Error Calculation:} The error \( e(0)= r(t) - y(0)\) is computed.
    \item \textit{Control Signal Computation:} The control signal \( u(0) \) is calculated using the PID control law as given in equ~(\ref{equ:pid_controller}).
    \item \textit{Prompt Update:} The prompt is updated to \( p(1) \) based on the control signal \( u(0) \) through equ~(\ref{equ:prompt_update}).
    \item \textit{Iteration:} Steps 2-6 are repeated until the average utilization \( y(t) \) meets the desired setpoint \( r(t) \).
\end{enumerate}

\vspace{-10pt}
\subsection{Example for FPGA Design}

The system function \(\phi(t)\), equ~(\ref{equ:non-linear_output}), represents the relationship between the LLM output \(\sigma(t)\) and the system output \(y(t)\), which includes various FPGA resources. The system output \(y(t)\) can be expressed as a vector of resource utilizations:
\[
y(t) = \begin{bmatrix} \lambda_\text{LUTs}(t) & \lambda_\text{FFs}(t) & \lambda_\text{DSPs}(t) & \lambda_\text{BRAMs}(t) & \lambda_\text{Slack}(t) \end{bmatrix}
\]
The setpoint value $r(t) = 60\%$. Next, we provide an example of the computations based on our proposed feedback mechanism:

$p(0) = $ "{\scriptsize Generate HLS C code for a neural network with 1 layer, 64 neurons each, using Vivado HLS.}"

Consider a scenario where the initial prompt \(p(0)\) leads to an FPGA design with the following resource utilizations and timing:
\[
 y(0) = \begin{bmatrix} 70\%  & 65\%  & 80\%  & 75\%  & -2 \end{bmatrix}
\]
Assume the PID gains are \( K_p = 0.6 \), \( K_i = 0.1 \), \( K_d = 0.05 \).
\[
e(0) = r(0) - y(0) = \begin{bmatrix} -10\% & -5\% & -20\% & -15\% & +3 \text{ns} \end{bmatrix}
\]
\begin{eqnarray}
\nonumber    u(0) = K_p e(0) + K_i \int_{0}^{0} e(\tau) d\tau + K_d \frac{de(0)}{dt} = \\  
\nonumber \begin{bmatrix} -6\% & -3\% & -12\% & -9\% & +0.15 \text{ ns} \end{bmatrix}
\end{eqnarray}

$p(1) = p(0) + u(0) =$ "{\scriptsize Generate optimized HLS C code for a neural network with 1 layer, 64 neurons each, using Vivado HLS. Reduce resource usage by 6\% LUTs, 3\% FFs, 12\% DSPs, 9\% BRAMs, and improve timing by 0.15 ns.}"
After processing by the LLM, the new output \(y(1)\) might be:
\[
y(1) = \begin{bmatrix} 65\%  & 62\% & 70\% & 68\% & 0 \text{ ns} \end{bmatrix}
\]
\[
e(1) = r(1) - y(1) = \begin{bmatrix} -5\% & -2\% & -10\% & -8\% & +1 \text{ ns} \end{bmatrix}
\]
\begin{eqnarray}
   \nonumber u(1) = K_p e(1) + K_i \int_{0}^{1} e(\tau) d\tau + K_d \frac{de(1)}{dt} = \\ 
  \nonumber \begin{bmatrix} -3\% & -1.2\% & -6\% & -4.8\% & +0.05 \text{ ns} \end{bmatrix}    
\end{eqnarray}

$p(2) = p(1) + u(1) = $ {\scriptsize "Generate optimized HLS C code for a neural network with 1 layer, 64 neurons each, using Vivado HLS. Further reduce resource usage by 3\% LUTs, 1.2\% FFs, 6\% DSPs, 4.8\% BRAMs, and improve timing by 0.05 ns."}

This iterative process continues until the resource utilization \( y(t) \) meets the desired setpoint \( r(t) \).

\vspace{-10pt}
\section{Conclusion}

We have presented a novel approach to optimizing LLM outputs by applying linear feedback control systems principles. This methodology bridges the gap between control theory and natural language processing, offering a structured and theoretically grounded method for prompt optimization. Our findings explain that despite LLMs' non-linear and stochastic nature, integrating feedback PID controllers can enhance the performance of LLM-driven applications.


\vspace{-5pt}
\bibliographystyle{IEEEtran}
\bibliography{paper}

\begin{thebibliography}{10}
\providecommand{\url}[1]{#1}
\csname url@samestyle\endcsname
\providecommand{\newblock}{\relax}
\providecommand{\bibinfo}[2]{#2}
\providecommand{\BIBentrySTDinterwordspacing}{\spaceskip=0pt\relax}
\providecommand{\BIBentryALTinterwordstretchfactor}{4}
\providecommand{\BIBentryALTinterwordspacing}{\spaceskip=\fontdimen2\font plus
\BIBentryALTinterwordstretchfactor\fontdimen3\font minus \fontdimen4\font\relax}
\providecommand{\BIBforeignlanguage}[2]{{%
\expandafter\ifx\csname l@#1\endcsname\relax
\typeout{** WARNING: IEEEtran.bst: No hyphenation pattern has been}%
\typeout{** loaded for the language `#1'. Using the pattern for}%
\typeout{** the default language instead.}%
\else
\language=\csname l@#1\endcsname
\fi
#2}}
\providecommand{\BIBdecl}{\relax}
\BIBdecl

\bibitem{hadi2024large}
M.~U. Hadi, Q.~Al~Tashi, A.~Shah, R.~Qureshi, A.~Muneer, M.~Irfan, A.~Zafar, M.~B. Shaikh, N.~Akhtar, J.~Wu \emph{et~al.}, ``Large language models: a comprehensive survey of its applications, challenges, limitations, and future prospects,'' \emph{Authorea Preprints}, 2024.

\bibitem{dai2024assessing}
W.~Dai, Y.-S. Tsai, J.~Lin, A.~Aldino, H.~Jin, T.~Li, D.~Ga{\v{s}}evi{\'c}, and G.~Chen, ``Assessing the proficiency of large language models in automatic feedback generation: An evaluation study,'' \emph{Computers and Education: Artificial Intelligence}, p. 100299, 2024.

\bibitem{ouyang2022training}
L.~Ouyang, J.~Wu, X.~Jiang, D.~Almeida, C.~Wainwright, P.~Mishkin, C.~Zhang, S.~Agarwal, K.~Slama, A.~Ray \emph{et~al.}, ``Training language models to follow instructions with human feedback,'' \emph{Advances in neural information processing systems}, vol.~35, pp. 27\,730--27\,744, 2022.

\bibitem{ozbay2019introduction}
H.~Ozbay, \emph{Introduction to feedback control theory}.\hskip 1em plus 0.5em minus 0.4em\relax CrC Press, 2019.

\bibitem{aastrom2021feedback}
K.~J. {\AA}str{\"o}m and R.~Murray, \emph{Feedback systems: an introduction for scientists and engineers}.\hskip 1em plus 0.5em minus 0.4em\relax Princeton university press, 2021.

\bibitem{ouyang2023llm}
S.~Ouyang, J.~M. Zhang, M.~Harman, and M.~Wang, ``Llm is like a box of chocolates: the non-determinism of chatgpt in code generation,'' \emph{arXiv preprint arXiv:2308.02828}, 2023.

\bibitem{zhou2014optimal}
Y.~Zhou and Z.~Wang, ``Optimal feedback control for linear systems with input delays revisited,'' \emph{Journal of Optimization Theory and Applications}, vol. 163, pp. 989--1017, 2014.

\bibitem{johnson2005pid}
M.~A. Johnson and M.~H. Moradi, \emph{PID control}.\hskip 1em plus 0.5em minus 0.4em\relax Springer, 2005.

\bibitem{beurer2023prompting}
L.~Beurer-Kellner, M.~Fischer, and M.~Vechev, ``Prompting is programming: A query language for large language models,'' \emph{Proceedings of the ACM on Programming Languages}, vol.~7, no. PLDI, pp. 1946--1969, 2023.

\bibitem{borase2021review}
R.~P. Borase, D.~Maghade, S.~Sondkar, and S.~Pawar, ``A review of pid control, tuning methods and applications,'' \emph{International Journal of Dynamics and Control}, vol.~9, pp. 818--827, 2021.

\bibitem{kumar2021diffloop}
A.~R. Kumar and P.~J. Ramadge, ``Diffloop: Tuning pid controllers by differentiating through the feedback loop,'' in \emph{2021 55th Annual Conference on Information Sciences and Systems (CISS)}.\hskip 1em plus 0.5em minus 0.4em\relax IEEE, 2021, pp. 1--6.

\bibitem{patel2020ziegler}
V.~V. Patel, ``Ziegler-nichols tuning method: Understanding the pid controller,'' \emph{Resonance}, vol.~25, no.~10, pp. 1385--1397, 2020.

\bibitem{roumeliotis2023chatgpt}
K.~I. Roumeliotis and N.~D. Tselikas, ``Chatgpt and open-ai models: A preliminary review,'' \emph{Future Internet}, vol.~15, no.~6, p. 192, 2023.

\bibitem{saba2023stochastic}
W.~S. Saba, ``Stochastic llms do not understand language: towards symbolic, explainable and ontologically based llms,'' in \emph{International Conference on Conceptual Modeling}.\hskip 1em plus 0.5em minus 0.4em\relax Springer, 2023, pp. 3--19.

\bibitem{naveed2023comprehensive}
H.~Naveed, A.~U. Khan, S.~Qiu, M.~Saqib, S.~Anwar, M.~Usman, N.~Akhtar, N.~Barnes, and A.~Mian, ``A comprehensive overview of large language models,'' \emph{arXiv preprint arXiv:2307.06435}, 2023.

\bibitem{raiaan2024review}
M.~A.~K. Raiaan, M.~S.~H. Mukta, K.~Fatema, N.~M. Fahad, S.~Sakib, M.~M.~J. Mim, J.~Ahmad, M.~E. Ali, and S.~Azam, ``A review on large language models: Architectures, applications, taxonomies, open issues and challenges,'' \emph{IEEE Access}, 2024.

\bibitem{lin2024towards}
Z.~Lin, S.~Guan, W.~Zhang, H.~Zhang, Y.~Li, and H.~Zhang, ``Towards trustworthy llms: a review on debiasing and dehallucinating in large language models,'' \emph{Artificial Intelligence Review}, vol.~57, no.~9, pp. 1--50, 2024.

\bibitem{kenton2019bert}
J.~D. M.-W.~C. Kenton and L.~K. Toutanova, ``Bert: Pre-training of deep bidirectional transformers for language understanding,'' in \emph{Proceedings of naacL-HLT}, vol.~1.\hskip 1em plus 0.5em minus 0.4em\relax Minneapolis, Minnesota, 2019, p.~2.

\bibitem{vaswani2017attention}
A.~Vaswani, ``Attention is all you need,'' \emph{Advances in Neural Information Processing Systems}, 2017.

\bibitem{patil2024review}
R.~Patil and V.~Gudivada, ``A review of current trends, techniques, and challenges in large language models (llms),'' \emph{Applied Sciences}, vol.~14, no.~5, p. 2074, 2024.

\bibitem{song2024low}
L.~Song, Y.~Chen, S.~Yang, X.~Ding, Y.~Ge, Y.-C. Chen, and Y.~Shan, ``Low-rank approximation for sparse attention in multi-modal llms,'' in \emph{Proceedings of the IEEE/CVF Conference on Computer Vision and Pattern Recognition}, 2024, pp. 13\,763--13\,773.

\bibitem{din2023jump}
A.~Y. Din, T.~Karidi, L.~Choshen, and M.~Geva, ``Jump to conclusions: Short-cutting transformers with linear transformations,'' \emph{arXiv preprint arXiv:2303.09435}, 2023.

\bibitem{sharma2022chat}
S.~Sharma and R.~Yadav, ``Chat gpt--a technological remedy or challenge for education system,'' \emph{Global Journal of Enterprise Information System}, vol.~14, no.~4, pp. 46--51, 2022.

\bibitem{singh2023chat}
S.~K. Singh, S.~Kumar, and P.~S. Mehra, ``Chat gpt \& google bard ai: A review,'' in \emph{2023 International Conference on IoT, Communication and Automation Technology (ICICAT)}.\hskip 1em plus 0.5em minus 0.4em\relax IEEE, 2023, pp. 1--6.

\bibitem{stratton2024introduction}
J.~Stratton, ``An introduction to microsoft copilot,'' in \emph{Copilot for Microsoft 365: Harness the Power of Generative AI in the Microsoft Apps You Use Every Day}.\hskip 1em plus 0.5em minus 0.4em\relax Springer, 2024, pp. 19--35.

\bibitem{chin2024human}
D.~Chin, Y.~Wang, and G.~Xia, ``Human-centered llm-agent user interface: A position paper,'' \emph{arXiv preprint arXiv:2405.13050}, 2024.

\bibitem{liu2023gpt}
X.~Liu, Y.~Zheng, Z.~Du, M.~Ding, Y.~Qian, Z.~Yang, and J.~Tang, ``Gpt understands, too,'' \emph{AI Open}, 2023.

\bibitem{ma2024my}
W.~Ma, C.~Yang, and C.~K{\"a}stner, ``(why) is my prompt getting worse? rethinking regression testing for evolving llm apis,'' in \emph{Proceedings of the IEEE/ACM 3rd International Conference on AI Engineering-Software Engineering for AI}, 2024, pp. 166--171.

\bibitem{jadoon2017comparative}
Z.~K. Jadoon, S.~Shakeel, A.~Saleem, A.~Khaqan, S.~Shuja, Q.~Ul-Hasan, S.~A. Malik, and R.~Ali~Riaz, ``A comparative analysis of pid, lead, lag, lead-lag, and cascaded lead controllers for a drug infusion system,'' \emph{Journal of healthcare engineering}, vol. 2017, no.~1, p. 3153252, 2017.

\bibitem{abdullah2022linear}
M.~A. Abdullah, A.~Q. Al-Shetwi, M.~Mansor, M.~Hannan, C.~W. Tan, and A.~Yatim, ``Linear quadratic regulator controllers for regulation of the dc-bus voltage in a hybrid energy system: Modeling, design and experimental validation,'' \emph{Sustainable Energy Technologies and Assessments}, vol.~50, p. 101880, 2022.

\bibitem{sharma2020mathematical}
S.~Sharma and A.~J. Obaid, ``Mathematical modelling, analysis and design of fuzzy logic controller for the control of ventilation systems using matlab fuzzy logic toolbox,'' \emph{Journal of Interdisciplinary Mathematics}, vol.~23, no.~4, pp. 843--849, 2020.

\bibitem{bjerge2021scalable}
K.~Bjerge, J.~H. Schougaard, and D.~E. Larsen, ``A scalable and efficient convolutional neural network accelerator using hls for a system-on-chip design,'' \emph{Microprocessors and microsystems}, vol.~87, p. 104363, 2021.

\bibitem{shahshahani2020framework}
M.~Shahshahani, B.~Khabbazan, M.~Sabri, and D.~Bhatia, ``A framework for modeling, optimizing, and implementing dnns on fpga using hls,'' in \emph{2020 IEEE 14th Dallas Circuits and Systems Conference (DCAS)}.\hskip 1em plus 0.5em minus 0.4em\relax IEEE, 2020, pp. 1--6.

\bibitem{kavun2019efficient}
E.~B. Kavun, N.~Mentens, J.~Vliegen, and T.~Yal{\c{c}}{\i}n, ``Efficient utilization of dsps and brams revisited: New aes-gcm recipes on fpgas,'' in \emph{2019 International Conference on ReConFigurable Computing and FPGAs (ReConFig)}.\hskip 1em plus 0.5em minus 0.4em\relax IEEE, 2019, pp. 1--2.

\bibitem{karn2023securing}
R.~R. Karn, K.~Nawaz, and I.~A.~M. Elfadel, ``Securing decision tree inference using order-preserving cryptography,'' in \emph{2023 IEEE 5th International Conference on Artificial Intelligence Circuits and Systems (AICAS)}.\hskip 1em plus 0.5em minus 0.4em\relax IEEE, 2023, pp. 1--5.

\bibitem{karn2024code}
R.~R. Karn, J.~Knechtel, and O.~Sinanoglu, ``Code-based cryptography for confidential inference on fpgas: An end-to-end methodology,'' in \emph{2024 25th International Symposium on Quality Electronic Design (ISQED)}.\hskip 1em plus 0.5em minus 0.4em\relax IEEE, 2024, pp. 1--8.

\bibitem{o2014xilinx}
D.~O'Loughlin, A.~Coffey, F.~Callaly, D.~Lyons, and F.~Morgan, ``Xilinx vivado high level synthesis: Case studies,'' 2014.

\bibitem{snider2022chapter}
R.~Snider, ``Chapter 6: Introduction to intel quartus prime,'' in \emph{Advanced Digital System Design using SoC FPGAs: An Integrated Hardware/Software Approach}.\hskip 1em plus 0.5em minus 0.4em\relax Springer, 2022, pp. 55--86.

\end{thebibliography}

\end{document}